\documentclass[nohyperref]{article}

\usepackage{microtype}
\usepackage{graphicx}
\usepackage{subfigure}
\usepackage{booktabs} 

\usepackage{hyperref}

\usepackage[accepted]{icml2022_pods}

\usepackage{amsmath}
\usepackage{amssymb}
\usepackage{mathtools}
\usepackage{amsthm}

\usepackage[capitalize,noabbrev]{cleveref}

\newcommand{\tmop}[1]{\ensuremath{\operatorname{#1}}}

\newcommand{\tmtextbf}[1]{\text{{\bfseries{#1}}}}
\newcommand{\tmtextit}[1]{\text{{\itshape{#1}}}}

\newcommand{\tmfloatcontents}{}
\newlength{\tmfloatwidth}
\newcommand{\tmfloat}[5]{
  \renewcommand{\tmfloatcontents}{#4}
  \setlength{\tmfloatwidth}{\widthof{\tmfloatcontents}+1in}
  \ifthenelse{\equal{#2}{small}}
    {\setlength{\tmfloatwidth}{0.45\linewidth}}
    {\setlength{\tmfloatwidth}{\linewidth}}
  \begin{minipage}[#1]{\tmfloatwidth}
    \begin{center}
      \tmfloatcontents
      \captionof{#3}{#5}
    \end{center}
  \end{minipage}}

\theoremstyle{plain}

\theoremstyle{definition}

\theoremstyle{remark}

\usepackage[textsize=tiny]{todonotes}

\icmltitlerunning{Noisy Learning for Neural ODEs Acts as a Robustness Locus Widening}

\begin{document}

\twocolumn[
\icmltitle{Noisy Learning for Neural ODEs Acts as a Robustness Locus Widening}



\icmlsetsymbol{equal}{*}

\begin{icmlauthorlist}
\icmlauthor{Martin Gonzalez}{yyy}
\icmlauthor{Hatem Hajri}{yyy}
\icmlauthor{Loic Cantat}{yyy}
\icmlauthor{Mihaly Petreczky}{comp}
\end{icmlauthorlist}

\icmlaffiliation{yyy}{Institut de Recherche Technologique SystemX, Palaiseau, France}
\icmlaffiliation{comp}{Centre de Recherche en Informatique, Signal et Automatique de Lille, France}

\icmlcorrespondingauthor{Martin Gonzalez}{martin.gonzalez@irt-systemx.fr}

\icmlkeywords{Machine Learning, ICML}

\vskip 0.3in
]



\printAffiliationsAndNotice{}  

\begin{abstract}
We investigate the problems and challenges of evaluating the robustness of Differential Equation-based (DE) networks against synthetic distribution shifts. We propose a novel and simple accuracy metric which can be used to evaluate intrinsic robustness and to validate dataset corruption simulators. We also propose methodology recommendations, destined for evaluating the many faces of neural DEs' robustness and for comparing them with their discrete counterparts rigorously. We then use this criteria to evaluate a cheap data augmentation technique as a reliable way for demonstrating the natural robustness of neural ODEs against simulated image corruptions across multiple datasets.
\end{abstract}

\section{Introduction}\label{Intro}

Neural Ordinary Differential Equations (NODEs) {\cite{chen}}, conjoining
dynamical systems (DS) and machine learning (ML), have come to be a popular source of
interest, in particular for tackling generative problems and continuous-time
modeling, and seem to have a bright future among the ML
community. Nevertheless, many questions regarding their robustness have been
raised, creating a debate on whether these networks benefit from
natural robustness properties or if the latter are overestimated.

Likewise, the importance of \tmtextit{reliability} in real-world applications
with AI-driven decision-making in safety-critical systems have brought a lot
of attention to studying a model's behavior under \tmtextit{distribution
shifts}. Understanding the latter implies focusing on how feasible is a chosen model's domain generalization against the kinds of shifts that
may occur in real-world scenarios. The past few years have seen an emerging industry proposing
new and relevant shifted datasets for different actors and purposes. Numerous
benchmarks {\cite{Hend,mnistc,wilds,salehi2021unified}} addressing different
aspects of distribution shifts have come to light and the rigorous
analysis and evaluation of both models and benchmarks have
become increasingly important. Although they transfer poorly to real-world
shifted images, synthetic distribution shifts are a good starting point for
experimenting a new model's accuracy and robustness. For instance, in
{\cite{gilmer2019adversarial}} it is hypothesized that methods that incur into
vanishing gradients also show no improvement in Gaussian noise, a phenomenon
which they relate in a rigorous way to adversarial attacks. Corruption
robustness can be then seen as a \tmtextit{sanity check} to ensure that a proposed
adversarial defense method doesn't present gradient masking. Nevertheless, it
is important to separate accuracy improvements from robustness improvements
when interpreting the results and different metrics have been proposed for doing
so {\cite{hendrycks2021many,taori2020measuring}}. For common corruptions {\cite{Hend}}, the (un-normalized, unaveraged)
relative Corruption Error (rCE) is the difference\footnote{Precise definitions are recalled in \eqref{rce}, Appendix \ref{app}.} of the model's corrupted and
clean errors. As the very notion of a corruption is always relative to a clean
counterpart, we find that this metric has a particular weakness for simulated
corruptions as it doesn't take into account the following structural principle
underlying such corruptions: \textit{miss-classified clean images should result in
miss-classified simulated corruptions}. As such, the rCE answers questions like
"how much does the model decline under corruption inputs" but it doesn't
detect the corruption error contributions coming from clean
miss-classifications.

This brief account aims to lay initial ground on theoretical and
application-driven aspects, problems and methodology perspectives for
evaluating robustness of NODEs against synthetic distribution shifts. For this
purpose, we
\begin{enumerate}
  \item assess and highlight several properties of NODEs in connection to
  different robustness criteria, link them to specific aspects of real-world
  data features they may capture and determine general guidelines on when and
  how they can be compared to static networks or between them;
  
  \item introduce an intrinsic robustness metric $\mathcal{A}^{\tmop{rel}}_c$,
  well-suited for evaluating well-posedness of dataset corruption simulators,
  and capable of measuring a model's corruption accuracy more subtly than the
  rCE {\cite{Hend}};
  
  \item evaluate an easy-to-implement robustifying method for NODEs against
  corrupted images, leading us to conclude that NODEs are naturally more
  robust to several synthetic distribution shifts than their discrete
  counterparts and that noisy learning for NODEs acts, as expected, as a
  robustness locus widening.
\end{enumerate}
We aim to propose a baseline upon which to build step-wise incremental
implementations of robustifying methods for NODEs under such corruptions. It
is our hope that our proposed metric and methodology recommendations will be
helpful both when studying implicit nets robustness and when designing new
and more diverse corruption simulation algorithms and datasets.

\section{On evaluating common robustness for neural ODEs}

Evaluating robustness of NODEs is particularly challenging: their output is
computed via iterative optimization schemes and such test-time optimization
has shown to prevent the proper evaluation of established robustness methods
designed for static networks like {\textsc{AutoAttack}} in the adversarial
context {\cite{croce22}}. Additionally, comparing NODEs to chosen static
analogs has shown to disregard implicit assumptions (adaptive step-size
solvers, inexact backward pass computation) which pose methodological problems
preventing to formally compare them and ultimately incurs into falsifying the
results of many conducted experiments. We will concentrate on classification
tasks in this report.

\tmtextbf{Neural ODEs meet dynamical systems:} Denote $h_x$ a feature
extractor (FE) and $h_y$ a fully-connected classifier (FCC). The inference of
a NODE model is carried out by solving,for $\dot{z}$ denoting the time-derivative:
\begin{equation}
  \left\{\begin{array}{l}
    \dot{z} (t) = f (t, z (t), \theta (t), x)\\
    z (0) = h_x (x)\\
    \hat{y} (T_x) = h_y (z (T_x))
  \end{array}\right. t \in \mathcal{T}_x = [0, T_x] \label{nODE}
\end{equation}
Contrary to static architectures, $f$ formalizes the dynamics controlling a
\tmtextit{continuous-in-depth} model, $t \in \mathcal{T}_x$ being its depth
variable and the components in \eqref{nODE} traduce the following features:
the dependence on $h_x$ for $z (0)$ is traduced by input layer augmentation;
the dependence on $t$ for $f$ (resp. $\theta$) is traduced by depth-dependence
(resp. depth-variance\footnote{When $\theta$ is a constant function, we still
use the term depth-dependence.}) and is taken in practice as an augmentation
component {\cite{dupont2019augmented}}; the dependence on $x$ for $f$ (resp.
$T_x$) is traduced as data-control (resp. depth-adaptation) and can traduce
recurrent architectures. We refer to {\cite{massaroli2020dissecting}} for
details on these features and to {\cite{kidgphd}} for a clear comprehensive
introduction to neural DEs.

Several overlaps occur between ML and DS
modeling techniques and approaches which we now try to articulate to shed
light on their singular benefits. These distinctions will be the basis of our
methodology guidelines for evaluating and comparing general Neural
Differential Equation (NDE) models.

\tmtextbf{DS-inspired neural DEs}: These consist in manufacturing constraints
on a loss function or on the weight matrices inside the dynamics that would
enhance their robustness from a stability analysis point of view. Another
way of stating DS-inspired NDEs is to say that the ML focus comes
\tmtextbf{post-hoc} the DS focus: the trained architecture is supposed to have
benefit of theoretical properties at training or inference. We highlight
the fact that such approaches can serve different purposes:
{\cite{pal2022mixing,RN11749,djeumou2022taylor}} address mainly speed
problems while {\cite{RN11716,RN11745,huang}} addresses stability training
considerations {\cite{ijcai2019}} for NODEs e.g. using steady-states, Lyapunov
equilibrium points. This usually is an idealized analysis made upon
an idealized ML architecture. A common problem would be to neglect the
numerical errors that come to hand while training, which were absent from
classical discrete neural networks. First, in {\cite{ott2020}} it is shown
that there exists a critical step size only beyond which the training yields a
valid ODE vector field. Thus, for instance, the system theoretic formulation
of the Picard-Lindel{\"o}f theorem, used for ensuring non intersecting
trajectories, effectively applies only if such condition is met.
\tmtextit{Methodology point}: ensure that the discretization resulting from
the numerical solver's execution preserves formalized DS properties. A first
characterization of robustness for NODEs is then met.

\tmtextbf{DS-based neural DEs}: These consist in formalizing NDE
architectures as analogs of system theoretic paradigms such as a full use of
the components of \eqref{nODE} but also formalizing neural CDEs, SDEs and PDEs
{\cite{kidgphd,fermanian2021framing,xu2022infinitely,li2021fourier}}. Here,
the ML focus comes \tmtextbf{ante-hoc} the DS focus: the formalized
architecture is supposed to be endowed with structural characteristics both at
training and inference. This approach is more involved than the previous one as
it needs to have at hand simultaneously an adapted, analytically proven,
adjoint method analog or generalization, and a non empty choice of adapted
numerical solvers. For instance, Kidger {\cite{kidgphd}} adapted the analytic
adjoint method for neural CDEs and neural SDEs while developing for the latter
an algebraically reversible \tmtextit{Heun method} SDE solver. While NODEs
have served as inspiration for constructing many discrete neural architectures
by formalizing in continuous-time an ODE and discretizing it, the difficulty
has been to create continuous time analogs for discrete neural components. For
instance, crafting a stateful batch normalization (BN) layer has been recently
reflected in the NODE formulation {\cite{que2021}} as a generalized ODE. On
the contrary, in {\cite{huang}} ResNets have BN layers, NODEs have group
normalization (GN) layers and the length of the skip connection does not
coincide between the compared architectures and in {\cite{xu2022infinitely}},
although testing NODEs against corruptions, a mix between deterministic and
stochastic methods may weaken their claims. \tmtextit{Methodology point}:
ensure that the chosen NDE architecture identifies in a clear manner all
arguments of the function passed to the DE solver, determine if the latter is
an exact or an approximate solver; distinguish stochastic and deterministic
architectures; comparing NDEs and discrete architectures should be
mathematically justified by an explicit end-to-end discretization scheme,
taking into account the nature of the $h_x$ and $h_y$ layers and identifying,
for instance, NODE blocks and weight-tied residual blocks. This
constitutes a second robustness characterization.

\tmtextbf{DS-destined neural DEs}: These consist in manufacturing NDEs that
incorporate known modeling physical constraints. Here, the ML focus is
\tmtextbf{ad-hoc} to the DS focus: the proposed architecture is supposed to
capture \tmtextit{intrinsically} the dynamics (e.g. Lagrangians, Hamiltonians)
of the studied phenomenon and NDEs \tmtextit{specify} DEs
{\cite{Zhong2020Symplectic}}. This is different, though somehow related, to
\tmtextit{physics-informed} NNs {\cite{karniadakis2021physics}}, which aim to
obtain solutions through NNs to \tmtextit{pre-specified} DEs for which
traditional solvers are computationally expensive. Well-defined NDEs may not
effectively capture continuous-time inductive biases if the used numerical ODE
methods have too low order of convergence while high-order methods need for
fast and exact gradient computations
{\cite{matsubara2021symplectic,djeumou2022taylor}}. \tmtextit{Methodology
point}: conduct NDE-oriented numerical \tmtextit{convergence} tests,
presenting a non-trivial difference between the ML-based {\cite{NIPS2007}} and
the round-off {\cite{chaitin1996lectures}} numerical errors, such as proposed
in {\cite{krishnapriyan2022learning}} to check if the implemented model
successfully learned meaningful continuous dynamics. We then find a third
robustness characterization for such networks.

Leveraging well-studied mathematical approaches for stability, robustness and
resilience\footnote{For instance, NODEs with depth-adaptation should be
evaluate their recovery rate in time \tmtextit{after} an input perturbation.} into the continuous-time ML community can prove to be very
advantageous and the above robustness properties, while already being studied jointly in the cited works, call for clear distinctions
between such notions. Their formal analysis will be the subject of an extended
version of the present report, complementary to the system-theoretic approach which is being conducted in parallel  \cite{gonzalez2022realization}. 

\tmtextbf{Intrinsic robustness metrics:} We now define the metric $\mathcal{A}^{\tmop{rel}}_c$ mentioned in the introduction of this report. Let $C$ be a
set of simulated corruptions $c = (\tilde{c}, s)$, where $\tilde{c}$ is a corruption
label and $s$ is a severity level. We make the assumption that for each $c \in
C$ one can generate (at least) one corruption simulation $x_c$ of a clean
image $x$. We denote $y$ the true label of $x$, $y_{\tmop{cl}}$ the model's
prediction on $x$ and $y_c$ the model's prediction on $x_c$. Let $N$ be the
size of the dataset. Define $M = \sum^N_{i = 1} \mathbb{I}_{\{y^i_{\tmop{cl}}
= y^i \}}$, $\mathcal{A}_{\tmop{cl}} = M / N,$ and for $c \in C$,
$$
\mathcal{A}_c=\frac{1}{N}\sum^N_{i = 1} \mathbb{I}_{\{y^i_c = y^i \}},
\quad \mathcal{A}^{\tmop{rel}}_c=\frac{1}{M}  \sum^N_{i = 1}
\mathbb{I}_{\{ y^i_c = y^i  \text{\& } y^i_{\tmop{cl}} = y^i \}}
$$
where $\mathbb{I}_{\{y_{\tmop{cl}} = y\}} = 1$ if $y_{\tmop{cl}} = y$ and 0
else. \textit{Methodology point}: the clean accuracy
$\mathcal{A}_{\tmop{cl}}$ is used to save the model's parameters during
training; the absolute corruption accuracy $\mathcal{A}_c$ gives the model's
accuracy for a corruption $c$; the relative corruption accuracy
$\mathcal{A}^{\tmop{rel}}_c$ computes how many corrupted simulations $x_c$
were correctly classified among the correctly classified clean counterparts;
the positiveness of the rCE remains a relevant sanity check for debugging and
verifying that the above hypothesis is preserved by the corruption simulator. 
By leveraging the above-mentioned structural principle for simulated corruptions, this metric addresses more accurately (in the statistical
analysis sense) questions like "how do corruptions intrinsically behave for
this model".

Our experiments show that $\mathcal{A}^{\tmop{rel}}_c$ doesn't overestimate
the model's robustness and abnormal behavior\footnote{Such as unexpectedly observing $\mathcal{A}_c>\mathcal{A}^{\tmop{rel}}_c$.} implies that a selected corruption simulation is
ill-posed. On the other hand, the Relative mCE increasingly
underestimates it, as highly accurate models will see a greater proportion of
their miss-classified corruptions to come from miss-classified clean samples,
making $\mathcal{A}_c$ to decrease while clean accuracy increases, as shown in
Fig. 3 of {\cite{Hend}}. This phenomenon is confirmed in Figures 5--8 of
{\cite{mnistc}} where miss-classified clean images represent 12\% of the shown
examples and none of them incur into well-classified associated corruptions.

\section{Experiments}

\begin{table*}[ht]
 \caption{Mean $\mathcal{A}^{\tmop{rel}}_c$ (\%) on corrupted MNIST. A $>5\%$ difference of model's performance is colored in orange. The last
  block computes the improvement on $\mathcal{A}^{\tmop{rel}}_c$ for each
  model induced by noisy training w.r.t. clean training. The listed
  corruptions are 1: gaussian, 2: shot, 3: impulse, 4: defocus, 5: glass, 6:
  motion, 7: zoom, 8: snow, 9: frost, 10: fog, 11: brightness, 12: contrast,
  13: elastic\_transform, 14: pixelate, 15: jpeg\_compression. }
  \vskip 0.15in
  \begin{center}
    \includegraphics[width=0.9\textwidth]{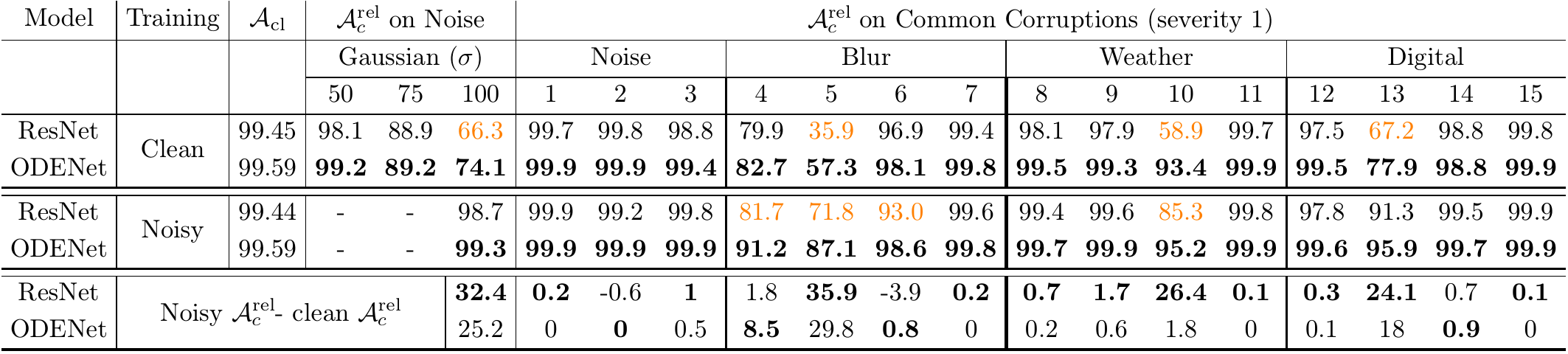}
  \end{center}
\end{table*}

\begin{table*}[ht]
 \caption{Mean $\mathcal{A}^{\tmop{rel}}_c$ (\%) at changes in corruption severity for
MNIST. At fixed $(c,s)\in C$, each block (in green) contains results for cleanly trained ResNet (upper-left), ODENet (lower-left) and their noisy counterparts (upper-right, lower-right) The listed corruptions are as in Table 1. Performance shifts are
colored in red. Corruptions where noisy training is not beneficial are colored
in blue.}
\vskip 0.15in
  \begin{center}
    \includegraphics[width=\textwidth]{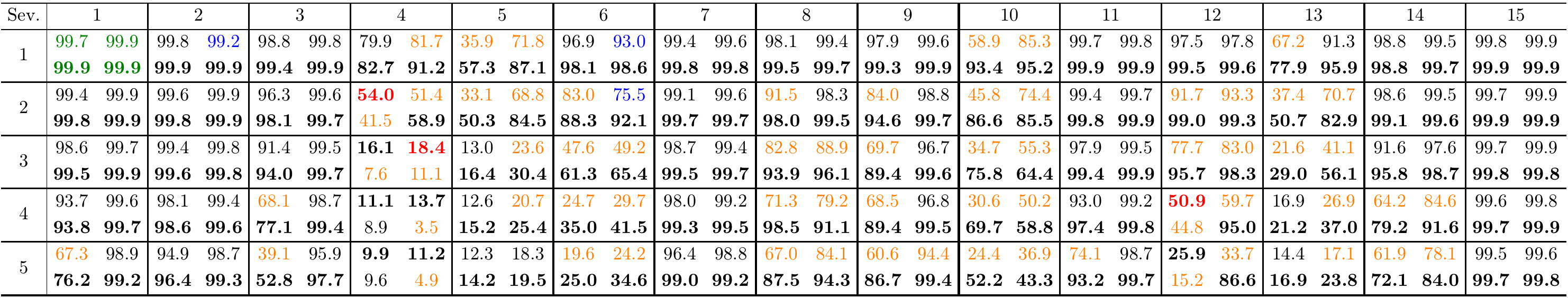}
  \end{center}
\end{table*}

We propose a minimalist, yet precise, comparative analysis on the robustness
of a simple NODE (ODENet) and its discrete counterpart (ResNet) against simulated image
corruptions. The chosen\footnote{Results of the remaining corrupted datasets and specifics on the chosen
architectures
are available in Appendix \ref{app}.} Models are
trained on MNIST with two methods:\tmtextit{ clean training} is done on clean-only
images; \tmtextit{noisy training} is conducted on a random combination of 50\%
of clean images and 50\% of images added Gaussian noise with randomly chosen
$\sigma \in \{ 50, 75, 100 \}$. We train each model on three different random
seeds, each trained model is then tested on 3 runs of corruption simulations
and only report the mean of the resulting 9 tests in Tables 1 \& 2. We report the mean of the 3 models clean accuracy $\mathcal{A}_{\tmop{cl}}$ used to save each model's parameters at which the rest of the
tests are conducted.

\tmtextbf{Results}: Table 1 shows that ODENet is consistently more robust
than ResNet and
that cleanly trained ODENet has less necessity of data augmentation to
achieve good performances than ResNet do, as seen in the last lines of Table 1, make us conclude that they are
naturally more robust than ResNet. This experimental
result is compatible with those appearing in the test-time adaptive models
literature {\cite{pmlr-v119-sun20b,wang2021tent}}. In light of the study in
{\cite{gilmer2019adversarial}} relating adversarial attacks as naturally
appearing in the scope of common corruptions, this result can also be thought
as a sanity check for determining that adversarial robustness for NODEs,
contrary to what is hypothesized in {\cite{huang}}, might \tmtextit{not} come
from obfuscated gradients. This fact seems to be further confirmed in
{\cite{chu2022improving}} although we have some reserves on their argument:
increasing the time horizon of a NODE should, in our opinion, rather be linked
to the model's resilience, roughly seen as the speed of
convergence \tmtextit{after} an input
perturbation, while robustness criteria usually focuses on the
distances and positions of inputs incurring on invariance of a model's prediction.
At increasing severity, as shown in Table 2, notice that, while ODENet is more robust
than ResNet on most corruptions, their decay of
robustness is bigger, shifting on some of the corruptions to
ResNets as the best model. Nonetheless, the same
shift occur at higher severity levels with noisy trained ODENets. Namely, for
defocus\_blur, on clean train mode, the shift was done at level 3 while at
noisy train mode it was only done at level 4. Analogously, for contrast
corruption, the shift at severity 4 on clean train mode was never reached on
noisy train mode. This sheds evidence to the fact that noisy training for
ODENets acts as a \tmtextit{robustness locus widening} i.e. that
the robustness neighborhood of data points $x$ become bigger with data
augmentation. Notice that these robustness neighborhoods are
\tmtextit{threat-model free}: they do not depend on the choice of
a norm ball as is commonly considered on
gradient-based defenses. Finally, noisy training made ResNet more vulnerable to
corruptions 2 and 6 at severity 1 but this vulnerability got corrected
at severity 3. This may suggest that partial information on the trade-off between accuracy and robustness may be captured
by a notion of model's deterioration \tmtextit{resilience} whose
rigorous study will be included in our upcoming extended study.

\section*{Acknowledgements}
The authors thank to each other for the fruitful conversations that led to the project for which this paper consists on a preliminary work. We would like to thank the reviewers for their comments. This work has been supported by the French government under the "France 2030” program, as part of the SystemX Technological Research Institute.

\bibliography{main}

\begin{thebibliography}{34}
\providecommand{\natexlab}[1]{#1}
\providecommand{\url}[1]{\texttt{#1}}
\expandafter\ifx\csname urlstyle\endcsname\relax
  \providecommand{\doi}[1]{doi: #1}\else
  \providecommand{\doi}{doi: \begingroup \urlstyle{rm}\Url}\fi

\bibitem[Bottou \& Bousquet(2007)Bottou and Bousquet]{NIPS2007}
Bottou, L. and Bousquet, O.
\newblock {The Tradeoffs of Large Scale Learning}.
\newblock In Platt, J., Koller, D., Singer, Y., and Roweis, S. (eds.),
  \emph{Advances in Neural Information Processing Systems}, volume~20. Curran
  Associates, Inc., 2007.
\newblock URL
  \url{https://proceedings.neurips.cc/paper/2007/file/0d3180d672e08b4c5312dcdafdf6ef36-Paper.pdf}.

\bibitem[Chaitin-Chatelin \& Frayss{\'e}(1996)Chaitin-Chatelin and
  Frayss{\'e}]{chaitin1996lectures}
Chaitin-Chatelin, F. and Frayss{\'e}, V.
\newblock \emph{{Lectures on finite precision computations}}.
\newblock SIAM, 1996.

\bibitem[Chen et~al.(2019)Chen, Rubanova, Bettencourt, and Duvenaud]{chen}
Chen, R. T.~Q., Rubanova, Y., Bettencourt, J., and Duvenaud, D.
\newblock {Neural Ordinary Differential Equations}.
\newblock In \emph{Neural Information Processing Systems (NeurIPS)}, 2019.
\newblock URL \url{http://arxiv.org/abs/1806.07366}.

\bibitem[Chu et~al.(2022)Chu, Wei, Lu, and Zhao]{chu2022improving}
Chu, H., Wei, S., Lu, Q., and Zhao, Y.
\newblock {Improving Neural ODEs via Knowledge Distillation}.
\newblock \emph{arXiv preprint arXiv:2203.05103}, 2022.

\bibitem[Croce et~al.(2022)Croce, Gowal, Brunner, Shelhamer, Hein, and
  Cemgil]{croce22}
Croce, F., Gowal, S., Brunner, T., Shelhamer, E., Hein, M., and Cemgil, T.
\newblock {Evaluating the Adversarial Robustness of Adaptive Test-time
  Defenses}.
\newblock \emph{arXiv preprint arXiv:2202.13711}, 2022.

\bibitem[Djeumou et~al.(2022)Djeumou, Neary, Goubault, Putot, and
  Topcu]{djeumou2022taylor}
Djeumou, F., Neary, C., Goubault, E., Putot, S., and Topcu, U.
\newblock {Taylor-Lagrange Neural Ordinary Differential Equations: Toward Fast
  Training and Evaluation of Neural ODEs}.
\newblock \emph{arXiv preprint arXiv:2201.05715}, 2022.

\bibitem[Dupont et~al.(2019)Dupont, Doucet, and Teh]{dupont2019augmented}
Dupont, E., Doucet, A., and Teh, Y.~W.
\newblock {Augmented Neural ODEs}, 2019.

\bibitem[Fermanian et~al.(2021)Fermanian, Marion, Vert, and
  Biau]{fermanian2021framing}
Fermanian, A., Marion, P., Vert, J.-P., and Biau, G.
\newblock {Framing RNN as a Kernel Method: A Neural ODE Approach}.
\newblock \emph{Advances in Neural Information Processing Systems}, 34, 2021.

\bibitem[Gilmer et~al.(2019)Gilmer, Ford, Carlini, and
  Cubuk]{gilmer2019adversarial}
Gilmer, J., Ford, N., Carlini, N., and Cubuk, E.
\newblock {Adversarial Examples are a Natural Consequence of Test Error in
  Noise}.
\newblock In \emph{International Conference on Machine Learning}, pp.\
  2280--2289. PMLR, 2019.

\bibitem[Gonzalez et~al.(2022)Gonzalez, Defourneau, Hajri, and
  Petreczky]{gonzalez2022realization}
Gonzalez, M., Defourneau, T., Hajri, H., and Petreczky, M.
\newblock {Realization Theory Of Recurrent Neural ODEs Using Polynomial System
  Embeddings}.
\newblock \emph{arXiv preprint arXiv:2205.11989}, 2022.

\bibitem[Hendrycks \& Dietterich(2019)Hendrycks and Dietterich]{Hend}
Hendrycks, D. and Dietterich, T.
\newblock {Benchmarking Neural Network Robustness to Common Corruptions and
  Perturbations}.
\newblock \emph{Proceedings of the International Conference on Learning
  Representations}, 2019.

\bibitem[Hendrycks et~al.(2021)Hendrycks, Basart, Mu, Kadavath, Wang, Dorundo,
  Desai, Zhu, Parajuli, Guo, et~al.]{hendrycks2021many}
Hendrycks, D., Basart, S., Mu, N., Kadavath, S., Wang, F., Dorundo, E., Desai,
  R., Zhu, T., Parajuli, S., Guo, M., et~al.
\newblock {The Many Faces Of Robustness: A Critical Analysis Of
  Out-Of-Distribution Generalization}.
\newblock In \emph{Proceedings of the IEEE/CVF International Conference on
  Computer Vision}, pp.\  8340--8349, 2021.

\bibitem[Huang et~al.(2022)Huang, Yu, Zhang, Ma, and Yao]{huang}
Huang, Y., Yu, Y., Zhang, H., Ma, Y., and Yao, Y.
\newblock {Adversarial Robustness of Stabilized Neural ODE Might be from
  Obfuscated Gradients}.
\newblock In \emph{Proceedings of the 2nd Mathematical and Scientific Machine
  Learning Conference}, volume 145 of \emph{Proceedings of Machine Learning
  Research}, pp.\  497--515. PMLR, 2022.

\bibitem[Ivan et~al.(2022)Ivan, Aaron, and Yue]{RN11749}
Ivan, Aaron, and Yue, Y.
\newblock {LyaNet}: A {Lyapunov} framework for training neural {ODEs}.
\newblock \emph{arXiv pre-print server}, 2022.
\newblock \doi{None arxiv:2202.02526}.

\bibitem[Kang et~al.(2021)Kang, Song, Ding, and Tay]{RN11716}
Kang, Q., Song, Y., Ding, Q., and Tay, W.~P.
\newblock {Stable Neural ODE with Lyapunov-Stable Equilibrium Points for
  Defending Against Adversarial Attacks}.
\newblock \emph{Advances in Neural Information Processing Systems}, 34, 2021.

\bibitem[Karniadakis et~al.(2021)Karniadakis, Kevrekidis, Lu, Perdikaris, Wang,
  and Yang]{karniadakis2021physics}
Karniadakis, G.~E., Kevrekidis, I.~G., Lu, L., Perdikaris, P., Wang, S., and
  Yang, L.
\newblock Physics-informed machine learning.
\newblock \emph{Nature Reviews Physics}, 3\penalty0 (6):\penalty0 422--440,
  2021.

\bibitem[Kidger(2022)]{kidgphd}
Kidger, P.
\newblock \emph{{On Neural Differential Equations}}.
\newblock PhD thesis, Oxford University, 2022.

\bibitem[Koh et~al.(2021)Koh, Sagawa, Marklund, Xie, Zhang, Balsubramani, Hu,
  Yasunaga, Phillips, Gao, Lee, David, Stavness, Guo, Earnshaw, Haque, Beery,
  Leskovec, Kundaje, Pierson, Levine, Finn, and Liang]{wilds}
Koh, P.~W., Sagawa, S., Marklund, H., Xie, S.~M., Zhang, M., Balsubramani, A.,
  Hu, W., Yasunaga, M., Phillips, R.~L., Gao, I., Lee, T., David, E., Stavness,
  I., Guo, W., Earnshaw, B., Haque, I., Beery, S.~M., Leskovec, J., Kundaje,
  A., Pierson, E., Levine, S., Finn, C., and Liang, P.
\newblock {WILDS: A Benchmark of in-the-Wild Distribution Shifts}.
\newblock In Meila, M. and Zhang, T. (eds.), \emph{Proceedings of the 38th
  International Conference on Machine Learning}, volume 139 of
  \emph{Proceedings of Machine Learning Research}, pp.\  5637--5664. PMLR,
  18--24 Jul 2021.

\bibitem[Krishnapriyan et~al.(2022)Krishnapriyan, Queiruga, Erichson, and
  Mahoney]{krishnapriyan2022learning}
Krishnapriyan, A.~S., Queiruga, A.~F., Erichson, N.~B., and Mahoney, M.~W.
\newblock {Learning Continuous Models for Continuous Physics}.
\newblock \emph{arXiv preprint arXiv:2202.08494}, 2022.

\bibitem[Li et~al.(2019)Li, Yi, Zhou, and Zhang]{ijcai2019}
Li, P., Yi, J., Zhou, B., and Zhang, L.
\newblock {Improving the Robustness of Deep Neural Networks via Adversarial
  Training with Triplet Loss}.
\newblock In \emph{Proceedings of the Twenty-Eighth International Joint
  Conference on Artificial Intelligence, {IJCAI-19}}, pp.\  2909--2915.
  International Joint Conferences on Artificial Intelligence Organization, 7
  2019.
\newblock \doi{10.24963/ijcai.2019/403}.
\newblock URL \url{https://doi.org/10.24963/ijcai.2019/403}.

\bibitem[Li et~al.(2021)Li, Kovachki, Azizzadenesheli, liu, Bhattacharya,
  Stuart, and Anandkumar]{li2021fourier}
Li, Z., Kovachki, N.~B., Azizzadenesheli, K., liu, B., Bhattacharya, K.,
  Stuart, A., and Anandkumar, A.
\newblock {Fourier Neural Operator for Parametric Partial Differential
  Equations}.
\newblock In \emph{International Conference on Learning Representations}, 2021.

\bibitem[Massaroli et~al.(2020)Massaroli, Poli, Park, Yamashita, and
  Asama]{massaroli2020dissecting}
Massaroli, S., Poli, M., Park, J., Yamashita, A., and Asama, H.
\newblock {Dissecting Neural ODEs}.
\newblock \emph{arXiv preprint arXiv:2002.08071}, 2020.

\bibitem[Matsubara et~al.(2021)Matsubara, Miyatake, and
  Yaguchi]{matsubara2021symplectic}
Matsubara, T., Miyatake, Y., and Yaguchi, T.
\newblock {Symplectic Adjoint Method for Exact Gradient of Neural {ODE} with
  Minimal Memory}.
\newblock In Beygelzimer, A., Dauphin, Y., Liang, P., and Vaughan, J.~W.
  (eds.), \emph{Advances in Neural Information Processing Systems}, 2021.
\newblock URL \url{https://openreview.net/forum?id=46J_l-cpc1W}.

\bibitem[Mu \& Gilmer(2019)Mu and Gilmer]{mnistc}
Mu, N. and Gilmer, J.
\newblock {MNIST-C: A Robustness Benchmark for Computer Vision}.
\newblock In \emph{ICML Workshop on Uncertainty and Robustness in Deep
  Learning}, 2019.

\bibitem[Ott et~al.(2021)Ott, Katiyar, Hennig, and Tiemann]{ott2020}
Ott, K., Katiyar, P., Hennig, P., and Tiemann, M.
\newblock {ResNet After All: Neural ODEs and Their Numerical Solution}.
\newblock In \emph{International Conference on Learning Representations}, 2021.

\bibitem[Pal et~al.(2022)Pal, Edelman, and Rackauckas]{pal2022mixing}
Pal, A., Edelman, A., and Rackauckas, C.
\newblock {Mixing Implicit and Explicit Deep Learning with Skip DEQs and
  Infinite Time Neural ODEs (Continuous DEQs)}.
\newblock \emph{arXiv preprint arXiv:2201.12240}, 2022.

\bibitem[Queiruga et~al.(2021)Queiruga, Erichson, Hodgkinson, and
  Mahoney]{que2021}
Queiruga, A.~F., Erichson, N.~B., Hodgkinson, L., and Mahoney, M.~W.
\newblock {Stateful ODE-Nets using Basis Function Expansions}.
\newblock In Beygelzimer, A., Dauphin, Y., Liang, P., and Vaughan, J.~W.
  (eds.), \emph{Advances in Neural Information Processing Systems}, 2021.

\bibitem[Salehi et~al.(2021)Salehi, Mirzaei, Hendrycks, Li, Rohban, and
  Sabokrou]{salehi2021unified}
Salehi, M., Mirzaei, H., Hendrycks, D., Li, Y., Rohban, M.~H., and Sabokrou, M.
\newblock {A Unified Survey on Anomaly, Novelty, Open-Set, and
  Out-of-Distribution Detection: Solutions and Future Challenges}.
\newblock \emph{arXiv preprint arXiv:2110.14051}, 2021.

\bibitem[Sun et~al.(2020)Sun, Wang, Liu, Miller, Efros, and
  Hardt]{pmlr-v119-sun20b}
Sun, Y., Wang, X., Liu, Z., Miller, J., Efros, A., and Hardt, M.
\newblock {Test-Time Training with Self-Supervision for Generalization under
  Distribution Shifts}.
\newblock In III, H.~D. and Singh, A. (eds.), \emph{Proceedings of the 37th
  International Conference on Machine Learning}, volume 119 of
  \emph{Proceedings of Machine Learning Research}, pp.\  9229--9248. PMLR,
  13--18 Jul 2020.

\bibitem[Taori et~al.(2020)Taori, Dave, Shankar, Carlini, Recht, and
  Schmidt]{taori2020measuring}
Taori, R., Dave, A., Shankar, V., Carlini, N., Recht, B., and Schmidt, L.
\newblock {Measuring Robustness to Natural Distribution Shifts in Image
  Classification}.
\newblock \emph{Advances in Neural Information Processing Systems},
  33:\penalty0 18583--18599, 2020.

\bibitem[Wang et~al.(2021)Wang, Shelhamer, Liu, Olshausen, and
  Darrell]{wang2021tent}
Wang, D., Shelhamer, E., Liu, S., Olshausen, B., and Darrell, T.
\newblock {Tent: Fully Test-Time Adaptation by Entropy Minimization}.
\newblock In \emph{International Conference on Learning Representations}, 2021.
\newblock URL \url{https://openreview.net/forum?id=uXl3bZLkr3c}.

\bibitem[Xu et~al.(2022)Xu, Chen, Li, and Duvenaud]{xu2022infinitely}
Xu, W., Chen, R.~T., Li, X., and Duvenaud, D.
\newblock {Infinitely Deep Bayesian Neural Networks with Stochastic
  Differential Equations}.
\newblock In \emph{International Conference on Artificial Intelligence and
  Statistics}, pp.\  721--738. PMLR, 2022.

\bibitem[Yan et~al.(2020)Yan, Du, Tan, and Feng]{RN11745}
Yan, H., Du, J., Tan, V., and Feng, J.
\newblock {On Robustness of Neural Ordinary Differential Equations}.
\newblock In \emph{International Conference on Learning Representations}, 2020.

\bibitem[Zhong et~al.(2020)Zhong, Dey, and Chakraborty]{Zhong2020Symplectic}
Zhong, Y.~D., Dey, B., and Chakraborty, A.
\newblock {Symplectic ODE-Net: Learning Hamiltonian Dynamics with Control}.
\newblock In \emph{International Conference on Learning Representations}, 2020.
\newblock URL \url{https://openreview.net/forum?id=ryxmb1rKDS}.

\end{thebibliography}
\bibliographystyle{icml2022}

\newpage
\appendix
\onecolumn
\section{Detailed experiment results}
\label{app}

In this Appendix we give further details on the application of our methodology
to the presented experiments: the chosen corrupt simulation algorithm is shown
to be non-trivial along different tested datasets (no miss-classified clean
images incur into well-classified corrupted counterparts); we do not include
corruptions in our train or validation sets, networks share the same $h_x$ and
$h_y$ modules; weight-tied ResNet blocks correspond to discretized NODE
blocks. Since our model is not DS-destined, we do not conduct a numerical
convergence test for the chosen Euler method.

\subsection{Model specifications}

All our models share the same FE and FCC modules and the RM modules consist on
the same layers to which one either applies a residual connection (for ResNet)
or the \tmtextit{odeint} function (for ODENet). In order to favor our ability
to compare ResNets and ODENets, we fix the Euler method as our ODE numerical
solver at time range $[0, 1]$ with 0.1 time steps which corresponds to ten weight-tied residual blocks. Finally, we use Group
Normalization (GN) \ instead of Batch Normalization (BN) to ensure that the
dynamics of the RM module truly correspond to an autonomous NODE.

\begin{table}[ht]
  \caption{The FE $h_x$ and FCC $h_y$ modules are identical for all our ResNet
  and ODENet models for MNIST, SVHN and CIFAR. The arguments of Conv2d are
  in order: the input channel, output channel, kernel size, stride and
  padding. Conv2dTime ensures time-dependence of the convolution component.
  The two arguments of the Linear layer represents the input dimension and the
  output dimension of this fully-connected layer.}
   \vskip 0.15in
  \begin{center}
     \includegraphics[width=0.5\textwidth]{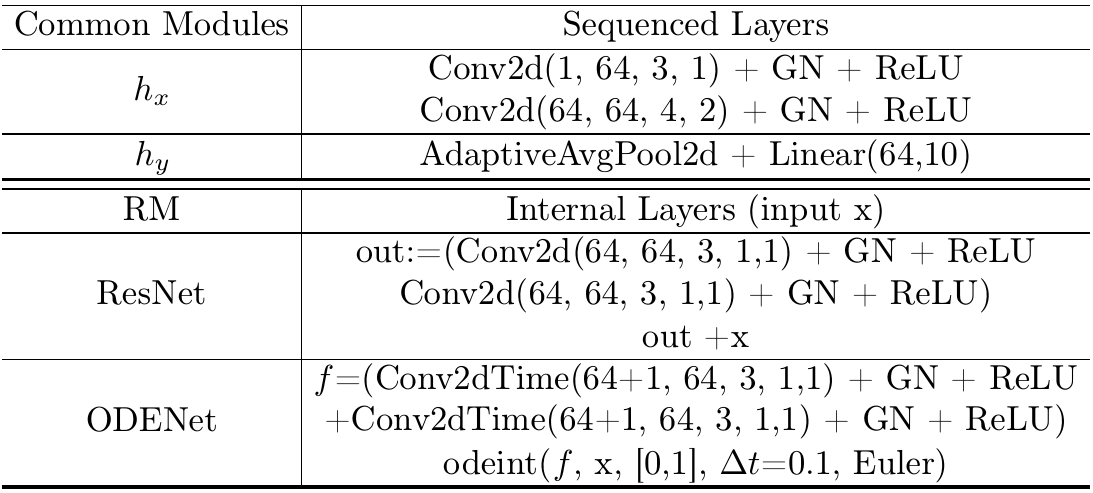}
  \end{center}
\end{table}

We train all our models for 100 epochs, learning rate 0.001, milestones [30,
60, 90]; decay 0.0005, $L_2$-penalty 0.2. Both models have around 142k
parameters.

While, in \eqref{nODE}, $f$ formalizes a single layer's dynamics, it usually
is taken in practice to be a composition of explicit functions that we pass to
the ODE solver (a block). Using BN as a block component holds mini-batch
information, which not only cannot be formalized as an autonomous ODE but may
lead to gradient explosion at back-propagation. When comparing ODENet and
ResNet blocks (without BN), one must ensure that each composite function for a
NODE block is both stateful (has an implicit dependence on the integration
time) and input-autonomous (does not present dependence or has leaked
information of the rest of the samples) either at training, validation or
testing. We use the ReLU function for practical purposes (increased
performance) after checking that models trained with fully differentiable
functions present the same behavior. We avoid taking into account in our
analysis neither neural SDEs, which can genuinely be seen as continuous-time
analog of noise injection robustifying methods, but whose inference is not
deterministic and which do not take into account stateful BN layers, as they
induce depth-varying architectures which do not have a clear discrete
counterpart upon which one could establish a comparative analysis. Choosing
the good basis function method for the latter to achieve competitive results
{\cite{que2021}} was a long effort, according to the authors, and hasn't yet
come close to state-of-the-art clean accuracy performances. In addition, BN
which has been shown to be crucial to achieve state-of-the-art robustness
performances, has also been shown to be a source of adversarial vulnerability
and it is unclear if their stateful counterpart from {\cite{que2021}} will
present or not this same behavior. Finally, our discretized NODE block is
formulated in terms of weight-tied ResNet blocks and matches the function to
which the residual skip connection is added with the one sent to the ODE
solver and does not match to the total length of convolution blocks present in
the architecture. For instance, appending 10 independent identical residual
convolutional blocks does not correspond to passing a single convolutional
block through a ODE solved with fixed 10 Euler steps as the state space of the
NODE is controlled by only one set of convolutional block parameters.

\subsection{Further Simulation Corrupted Dataset Experiments}

We use several datasets and their synthetic corruptions by using the same
simulation algorithm and selecting corruptions that make sense on all
datasets. Noisy training is done with randomly added noise with $\sigma \in \{
10, 15, 25 \}$ for the SVHN dataset, and with $\sigma \in \{ 10, 15, 20 \}$
for the CIFAR10 dataset.  Our experiment's relative perturbed
accuracy is given in Tables 4 and 5 for SVHN\footnote{The results for CIFAR showing no novel or different behaviour than the one conducted for SVHN other than an overall drop in accuracy, we do not present them in this preliminary report.}.

\begin{table}[ht]
  \caption{Mean $\mathcal{A}^{\tmop{rel}}_c$ (\%) on corrupted SVHN images for
  ResNet and ODENet. The listed corruptions are as in Table 1. The last block computes the improvement on performance
  for each model induced by noisy training w.r.t. clean training. Corruptions where noisy training is not beneficial are colored in blue and a $>5\%$ difference of model's performance is colored in orange.}
  \vskip 0.15in
  \begin{center}
    \includegraphics[width=0.9\textwidth]{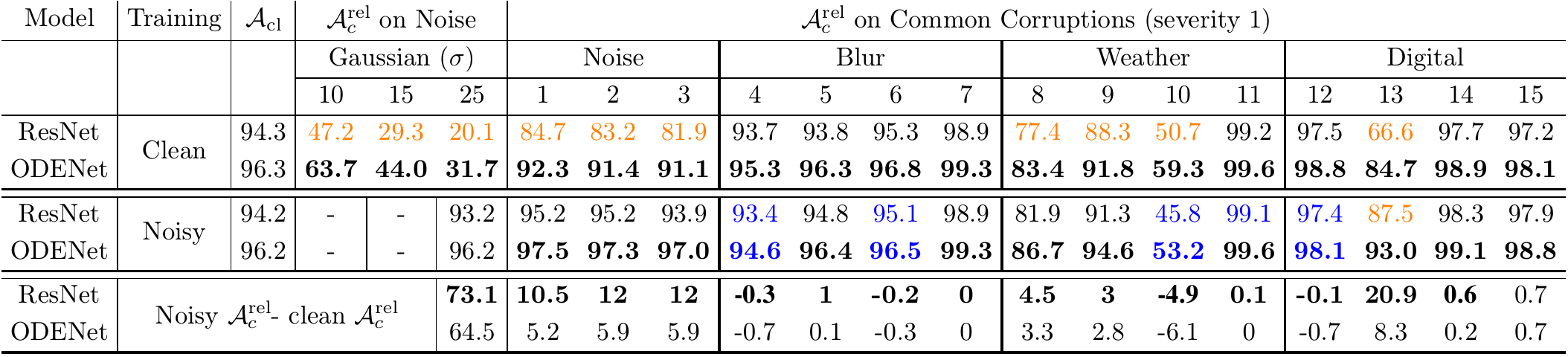}
  \end{center}
\end{table}

\begin{table}[ht]
  \caption{Mean $\mathcal{A}^{\tmop{rel}}_c$ (\%) at changes in severity. The
  listed corruptions are as in Table 1. Red color means a shift of best
  model's accuracy w.r.t. previous severity value. Corruptions where noisy training is not beneficial are colored in blue.}
   \vskip 0.15in
  \begin{center}
   \includegraphics[width=0.75\textwidth]{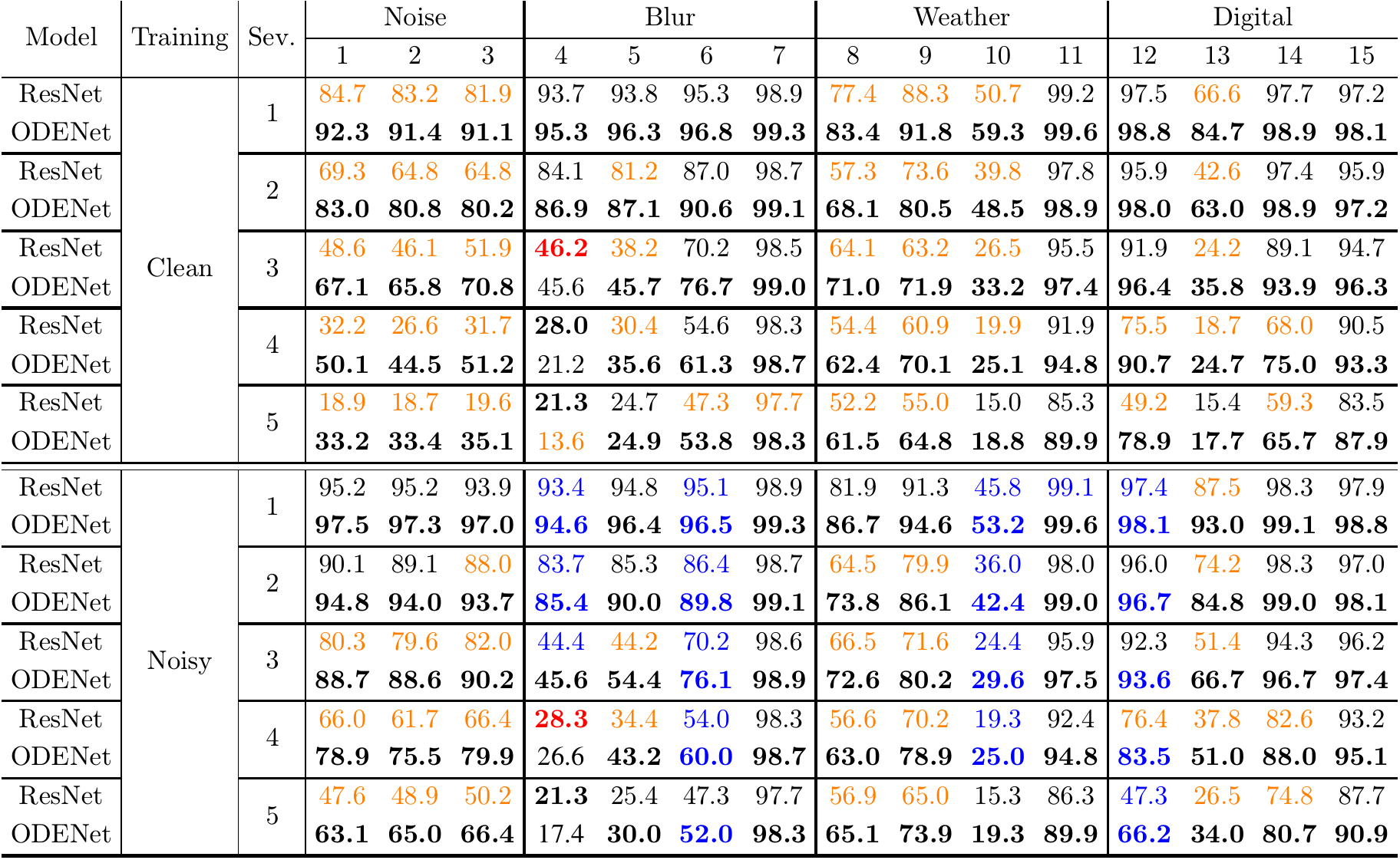}
  \end{center}
\end{table}

\textbf{Results:} ODENet is consistently more robust than
ResNet and the latter benefited more than ODENet from noisy training. This confirms our MNIST conclusions on the natural robustness of ODENet. Notice that noisy training makes the models slightly more vulnerable for some
corruptions. At severity changes, noisy training acts here again as a
robustness locus widening. Interestingly, for those corruptions where noisy training made the model more
vulnerable at severity 1, noisy trained models see their vulnerability remain only on half of those corruptions at severity 5. Finally, as shown in orange, it seems that noisy training makes ODENet more resilient to corruption deterioration than it does to ResNet.

\subsection{Further thoughts on corruption simulators and their metrics}

\tmtextbf{Baseline Normalization}: Usual corruption metrics such as the
$\tmop{mCE}^f$ and the relative $\tmop{mCE}^f$ for a model $f$ are computed with respect to a
baseline (e.g. AlexNet) error in order to normalize it over those corruptions
that are known to be particularly challenging. For $(c, s)\in C$, they are the average over all 15 corruptions labels $c$ of the clean and corrupted top-1 error rates (averaged first across all 5 severity levels):
\begin{equation}\label{rce}
    \text{CE}_c^f = \bigg(\sum_{s=1}^5 E_{s,c}^f\bigg)\bigg/\bigg(\sum_{s=1}^5 E_{s,c}^\text{AlexNet}\bigg), \qquad \text{rCE}_c^f = \bigg(\sum_{s=1}^5 E_{s,c}^f - E^f_\text{clean}\bigg)\bigg/\bigg(\sum_{s=1}^5 E_{s,c}^\text{AlexNet} - E^\text{AlexNet}_\text{clean}\bigg)
\end{equation}
We do not normalize our metrics
for two reasons. First, the $\tmop{mCE}^f$ was proposed as an attempt to unify
robustness bench-marking under a unique number across many models, which we do
not do here. But most importantly, it has been shown that feature aggregating
networks and deeper nets markedly enhance robustness. Thus, as NODEs consist
on a paradigm shift of the notion of depth, which becomes adaptive even while
testing, we find that testing such network against a baseline fixed depth
network such as AlexNet could be harmful for our comparative analysis. This is
somewhat concordant with the last recommendation in {\cite{croce22}} for
generating adversarial attacks for adaptive test-time defenses. We do not average our metrics across severity levels to analyse their behaviour at increasing severity.

\tmtextbf{Corruption simulators:} Instead of testing our models on static
corrupted datasets (MNIST-C, CIFAR10-C, ImageNet-C), we run the exact same
corruption
simulator\footnote{Available at \url{https://github.com/bethgelab/imagecorruptions}.} on the
datasets of all our experiments (and which is the same one used by the authors
that proposed the above-mentioned static corrupted datasets). For simplicity,
corruptions that were specially tailored for MNIST (such as zig-zag or canny edges)
that only make sense to be conducted on that dataset will be left out from our
comparative study, as well as fully formalizable corruptions (such as
rotations, translations..) that one can use as auxiliary data augmentation
techniques such as adversarial and S\&P noise augmentations. This should be
taken into account as a partial reproducibility issue: while the performance
of the considered models should decrease when tested on compressed JPEG
corrupted datasets such as ImageNet-C, the comparative results conducted in
this work do not show any qualitative distinction (although the overall
model's accuracies decrease). Also, our objective is not to propose an
architecture capable of achieving state-of-the-art robust performances in
either of the mentioned datasets. Our choice to fix a common corruption
simulator for different datasets is somehow a model-driven choice. It has been
shown that classification error patterns between robust models and those
coming from human perception are fundamentally different. As such, focusing on
understanding a model's behavior around simulated corruptions can be improved
by fixing one simulator and generating corruptions among different datasets.
This allows to release some part of the randomness of a generated simulation
and prevents different data augmentation techniques to guess a corruption
simulator's parameters, which in a sense can be seen as information leakage.
By using the corruption simulator as a white-box component of our generated
corruption dataset we hope this will promote better model's transferability to
some degree.


\end{document}